\documentclass[10pt]{article} 

\usepackage[preprint]{rlj}

\usepackage{amssymb}            
\usepackage{mathtools}         
\usepackage{mathrsfs}           
\usepackage{makecell}           
\usepackage{graphicx}          
\usepackage{subcaption}        
\usepackage[space]{grffile}     
\usepackage{url}               
\usepackage{lipsum}            
\usepackage{booktabs}
\usepackage{array}
\usepackage{subcaption}
\usepackage{pifont}             
\usepackage{xcolor}            
\usepackage{listings}           
\usepackage{svg}

\newcommand{\hide}[1]{}

\definecolor{codegreen}{rgb}{0,0.6,0}
\definecolor{codegray}{rgb}{0.01,0.01,0.01}
\definecolor{codepurple}{rgb}{0.58,0,0.82}
\definecolor{backcolour}{rgb}{0.95,0.95,0.92}

\lstdefinestyle{mystyle}{
    backgroundcolor=\color{backcolour},   
    commentstyle=\color{codegreen},
    keywordstyle=\color{codegray},
    numberstyle=\tiny\color{codegray},
    stringstyle=\color{codepurple},
    basicstyle=\ttfamily\footnotesize\color{codegray},
    breakatwhitespace=false,         
    breaklines=true,                 
    captionpos=b,                    
    keepspaces=true,                 
    numbers=none,                    
    numbersep=5pt,                  
    showspaces=false,                
    showstringspaces=false,
    showtabs=false,                  
    tabsize=2
}

\title{WOFOSTGym: A Crop Simulator for Learning Annual and Perennial Crop Management Strategies}

\setrunningtitle{WOFOSTGym}

\author{William Solow\textsuperscript{1}, Sandhya Saisubramanian\textsuperscript{1}, Alan Fern\textsuperscript{1}}

\emails{ \{soloww,sandhya.sai,alan.fern\}@oregonstate.edu}

\affiliations{
$^{1}$\textbf{Oregon State University, Corvallis, Oregon}\\
}

\begin{document}

\maketitle 

\begin{abstract}
We introduce WOFOSTGym, a novel crop simulation environment designed to train reinforcement learning (RL) agents to optimize agromanagement decisions for annual and perennial crops in single and multi-farm settings. Effective crop management requires optimizing yield and economic returns while minimizing environmental impact, a complex sequential decision-making problem well suited for RL. However, the lack of simulators for perennial crops in multi-farm contexts has hindered RL applications in this domain. Existing crop simulators also do not support multiple annual crops. WOFOSTGym addresses these gaps by supporting 23 annual crops and two perennial crops, enabling RL agents to learn diverse agromanagement strategies in multi-year, multi-crop, and multi-farm settings. Our simulator offers a suite of challenging tasks for learning under partial observability, non-Markovian dynamics, and delayed feedback. WOFOSTGym's standard RL interface allows researchers without agricultural expertise to explore a wide range of agromanagement problems. Our experiments demonstrate the learned behaviors across various crop varieties and soil types, highlighting WOFOSTGym's potential for advancing RL-driven decision support in agriculture.
\end{abstract}

\section{Introduction}
\label{sec:intro}
During a growing season, farmers face many decisions about how to optimally manage their crops to increase yield while reducing cost and environmental impact~\citep{javaid_understanding_2023}. For example, irrigation planning must account for constraints on water use, and optimal irrigation scheduling can improve crop yield~\citep{elliott_constraints_2014}. Motivated by the promising results of using reinforcement learning (RL) in other areas of precision agriculture, there is increasing interest from researchers and government agencies in applying RL to crop management decision problems in open-field agriculture, especially for perennial crops (e.g. pears, grapes)~\citep{astill_developing_2020,gautron_reinforcement_2022}. 

Agriculture presents key challenges for RL, making it a valuable testbed for research: (1) \emph{delayed feedback}---actions like fertilization affect yield only months later, complicating credit assignment; (2) \emph{sparse rewards}---since yield is only known at the episode's end, learning an optimal policy is difficult~\citep{vecerik_leveraging_2018}; and (3) \emph{partial observability}---many crop and soil states are unmeasurable or costly to obtain. While RL has been explored as a tool for optimizing open-field crop management decisions~\citep{wu_optimizing_2022, tao_optimizing_2023}, its real-world adoption is limited to controlled settings such as greenhouses~\citep{an_simulator-based_2021, wang_deep_2020} and crop monitoring~\citep{din_deep_2022, zhang_whole-field_2020}. We bridge this gap by presenting a simulator for annual and perennial crops in single and multi-farm settings.

Training RL agents in the real world to optimize agromanagement decisions is infeasible because growing seasons are too long, and unconstrained exploration can cause costly errors like crop death and soil degradation~\citep{tevenart_role_2021}. Similar challenges in robotics and autonomous driving have been addressed with high-fidelity simulators, enabling RL applications~\citep{kober_reinforcement_2013,kiran_deep_2022,dauner_navsim_2024,todorov_mujoco_2012}. While high-fidelity crop growth models (CGMs)~\citep{boote_potential_1996} offer an approach to testing crop management policies, they are \emph{not} designed to interact with RL algorithms and require substantial domain expertise. 

Existing agriculture simulators~\citep{gautron_gym-dssat_2022} only simulate the growth of a \emph{single} annual crop. They lack the functionality needed for perennial crop management as they do not capture crop growth across multiple years, including the dormant season~\citep{forcella_real-time_1998} (see Table~\ref{tab:crop_benchmarks}). 
Moreover, these simulators cannot be customized to study other crops or sites without domain knowledge of the underlying CGM and cannot learn joint agromanagement policies for multiple farms. Open-field agriculture problems are often modeled as a partially observable environment, but the current crop simulators do not allow for varying the number of hidden features for partial observability and do not support the creation of a wide range of agromanagement tasks across crop and soil types, which limit the scenarios that can be modeled~\citep{tao_optimizing_2023}.

\begin{figure}[t]
    \centering
    \includegraphics[width=0.97\linewidth]{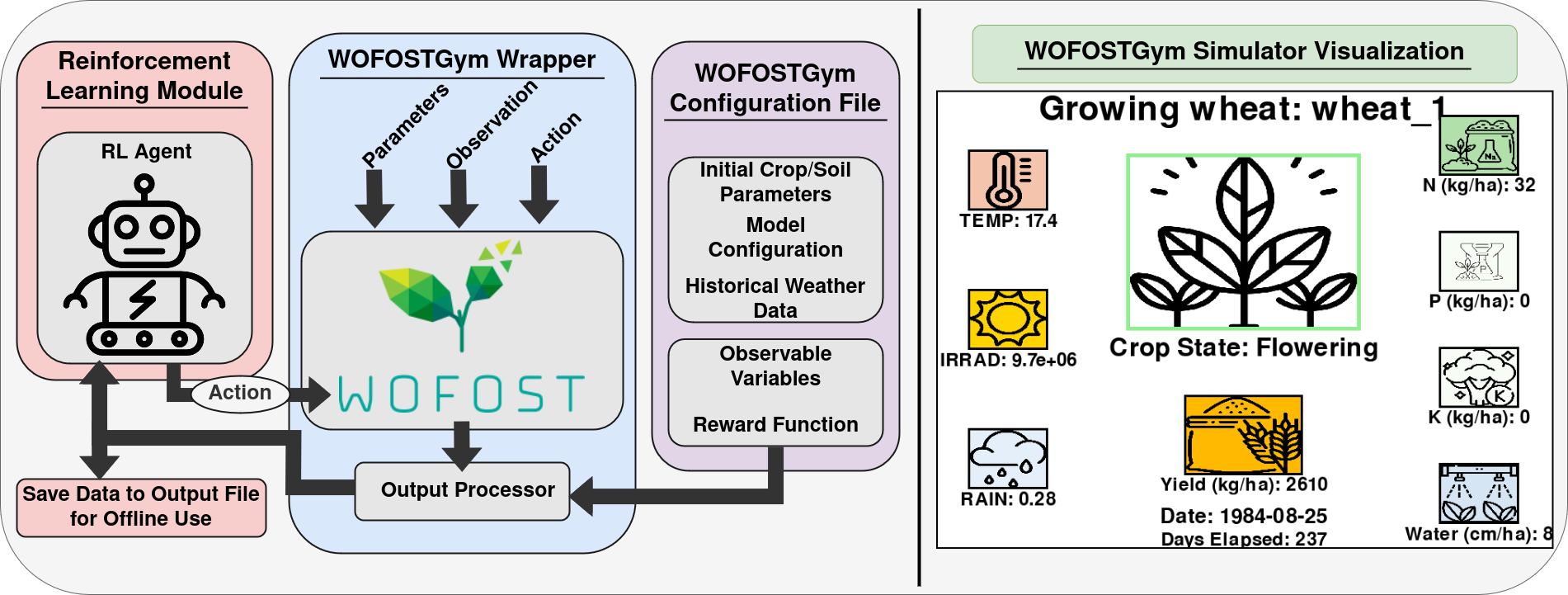}
    \caption{The structure and visualization of the WOFOSTGym simulator. WOFOSTGym provides an API around the WOFOST Crop Growth Model with a variety of environments to train RL agents and generate data. Well documented configuration files control crop and soil dynamics.}
    \label{fig:wofost_structure}
\end{figure}

We present WOFOSTGym (see Figure~\ref{fig:wofost_structure}), a crop simulator for learning annual and perennial crop management strategies across single and multiple farms.
WOFOSTGym is built on the WOFOST CGM~\citep{van_diepen_wofost_1989} to model the growth of perennial crops, and includes parameter sets for 23 annual crops and two perennial crops. Each crop contains between one and ten varieties. As a step towards high-fidelity modeling of perennial crop growth, we employ a Bayesian Optimization based method to calibrate CGM parameters to increase the fidelity of the phenological model for 32 grape cultivars. To make WOFOSTGym accessible to RL researchers, we prioritize usability through extensive customization, seamless integration with standard RL algorithms, and a thorough documentation: \url{https://intelligent-reliable-autonomous-systems.github.io/WOFOSTGym-Docs/}.

Our experiments highlight scenarios in WOFOSTGym where standard RL algorithms and imitation learning (IL) agents achieve optimal performance, alongside more complex cases that remain difficult, underscoring opportunities for advancing learning approaches in agromanagement for both annual and perennial crops. We also design agromanagement decision-making tasks in WOFOSTGym that illustrate both the potential and challenges of applying RL to agriculture, positioning WOFOSTGym as a rigorous testbed for developing and evaluating new algorithms.

\section{Background and Related Work}
\label{sec:related}

This section briefly describes the POMDP framework used for modeling our problems, and an overview of existing crop simulators and crop growth models. 
\paragraph{Partially Observable Markov Decision Process} We formulate our agromanagement problems using the framework of partially observable Markov decision process (POMDP)~\citep{kaelbling_planning_1998}.
POMDPs are well-suited for open-field agriculture problems, since many crop and soil-related features that are essential for defining the system's full state cannot be directly observed~\citep{tao_optimizing_2023}. Formally, a POMDP is a tuple $M=\langle \mathcal{S}, \mathcal{A}, \mathcal{P}, \mathcal{R}, \Omega, \mathcal{O} \rangle$ where $\mathcal{S}$ is a set of states, $\mathcal{A}$ is a set of actions, $\mathcal{P}:\mathcal{S}\times \mathcal{A}\times\mathcal{S}\to[0,1]$ is the transition kernel, and $\mathcal{R}:\mathcal{S}\times\mathcal{A}\to\mathbb{R}$ is the reward function. $\Omega$ is the set of possible observations and $\mathcal{O}:\mathcal{S}\times\mathcal{A}\times\Omega\to[0,1]$ is the probability of obtaining observation $o$ when taking action $a$ in state $s$. A reward discount factor $\gamma$ determines the importance of immediate versus future rewards. The RL agent computes a policy $\pi:\Omega\times \mathcal{A}\to[0,1]$ that maximizes the expected sum of discounted rewards, $\mathbb{E}_{\rho^\pi}\left[\sum_{t=0}^T\gamma^t\mathcal{R}(s_t,a_t)\right]$, where $\rho^\pi$ is the distribution of states and actions induced by the policy $\pi$ and $T$ is the time horizon. 

\paragraph{RL for Crop Management}
Building on RL's success in robotics, autonomous driving, and healthcare, there is growing interest in applying RL to optimize crop yield~\citep{binas_reinforcement_2019}. While RL has proven effective in controlled greenhouse environments~\citep{an_simulator-based_2021}, its application in open-field agriculture remains limited due to reduced sensing capabilities and long growing seasons. \citet{tao_optimizing_2023} proposed an imitation learning approach to learn expert actions under partial observability, but it has not been tested in the real-world. To bridge this gap, several crop simulators have been developed. CropGym simulates winter wheat in a nitrogen-limited soil via a Gym wrapper around a CGM~\citep{overweg_cropgym_2021}. Gym-DSSAT focuses on maize growth optimization through fertilization and irrigation decisions~\citep{gautron_gym-dssat_2022}. CyclesGym, built around the Cycles CGM~\citep{kemanian_cycles_2022}, focuses on learning crop rotation strategies for annual crops but is limited to soybeans and maize, lacking support for perennial crop modeling~\citep{turchetta_learning_2022}. Table~\ref{tab:crop_benchmarks} summarizes the capabilities of different crop simulators.

Existing crop simulators support RL training for fertilization and irrigation but \emph{lack support for perennial crops}, a key research area~\citep{gautron_reinforcement_2022}. Additionally, customization is infeasible without expert knowledge of the underlying CGMs, since most CGMs are run through separate executables. In contrast, WOFOSTGym offers easy domain customization for RL researchers while providing high-fidelity parameters for 23 annual and two perennial crops, high-fidelity model parameters for grape phenology for 32 cultivars, and access to diverse soil types and weather patterns.

\begin{table}[t]
\begin{tabular}{l cccccccc }
\hline
Name & \makecell{Perennial Crop\\Support} & \makecell{Multiple Crops\\and Farms}& \makecell{Easily\\Customizable} & \makecell{Models Crop\\Sub-processes}  \\
\hline
CyclesGym & \textcolor{red}{\ding{55}} & \textcolor{red}{\ding{55}} & \textcolor{green}{\ding{51}}& \textcolor{green}{\ding{51}}\\
\hline
CropGym & \textcolor{red}{\ding{55}} & \textcolor{red}{\ding{55}} & \textcolor{red}{\ding{55}}& \textcolor{green}{\ding{51}} \\ 
\hline
gym-DSSAT & \textcolor{red}{\ding{55}} &\textcolor{red}{\ding{55}} & \textcolor{red}{\ding{55}}& \textcolor{green}{\ding{51}} \\
\hline
FarmGym & \textcolor{red}{\ding{55}} & \textcolor{red}{\ding{55}} & \textcolor{green}{\ding{51}} & \textcolor{red}{\ding{55}} \\
\hline
WOFOSTGym (Ours) & \textcolor{green}{\ding{51}} & \textcolor{green}{\ding{51}} & \textcolor{green}{\ding{51}} & \textcolor{green}{\ding{51}} \\
\hline
\end{tabular}
\caption{Comparison of available crop simulators based on four important desiderata for use with RL. A simulator is easily customizable if it does not require agriculture domain expertise to run different experiments. Modeling crop sub-processes (phenology, roots, stems, leaves, etc.) as it generally leads to a higher fidelity model.}
\label{tab:crop_benchmarks}
\end{table}

\paragraph{Crop Growth Models}
Crop Growth Models (CGMs) simulate the growth of crops in varying environments subject to different agromanagement decisions~\citep{jones_brief_2017}.
Examples of widely-used CGMs include WOFOST~\citep{de_wit_25_2019}, DSSAT~\citep{jones_dssat_2003}, APSIM~\citep{mccown_apsim_1996}, EPIC~\citep{cabelguenne_calibration_1990}, CropSyst~\citep{stockle_cropsyst_1994}, Cycles~\citep{kemanian_cycles_2022} and AquaCrop~\citep{andarzian_validation_2011}. None of the available CGMs support perennial crops. The relevant features of these CGMs are highlighted in the Appendix. 

Our simulator is built on WOFOST, a CGM that models annual crop growth subject to nutrient (nitrogen, phosphorus, and potassium) and water-limited conditions~\citep{van_diepen_wofost_1989}. We choose WOFOST CGM since it can model the growth of perennial crops with a high fidelity~\citep{bai_growth_2020,shi_yield_2022}. It also accounts for varying CO2 concentrations, making it valuable for climate-impacted agricultural research~\citep{gilardelli_sensitivity_2018}. Additionally, its modular design facilitates modifications to crop process models~\citep{de_wit_pcse_2024}, and its Python implementation enables seamless integration with OpenAI Gym~\citep{brockman_openai_2016}.

\section{WOFOSTGym}
\label{sec:wofostgym}
WOFOSTGym is built on the WOFOST CGM~\citep{van_diepen_wofost_1989} and interfaces with the OpenAI Gym API to create a high-fidelity and easy-to-use crop simulator for RL. Agromanagement decisions supported in
WOFOSTGym are: fertilizing, irrigating, planting, and harvesting. In the interest of clarity, we focus on fertilization and irrigation decisions in the rest of the paper, since these tasks are supported by all existing crop simulators. In these tasks, the agent must optimize fertilization and irrigation strategies that maximize the cumulative yield of a crop subject to a set of penalties or constraints over one or more growing seasons and across one or more farms. 

The rest of this section is organized as follows. We begin with an overview of the environment design. We then propose a model calibration method to fine-tune the model parameters of the WOFOST CGM to increase the fidelity as a step towards sim-to-real transfer~\citep{peng_sim--real_2018}.

\subsection{Environment Design}
A WOFOSTGym instance is defined by its Gym environment ID, the reward wrapper, and an agromanagement configuration file. WOFOSTGym contains 54 Gym environments that relate to annual and perennial crop simulation, single and multi-farm simulations, and six combinations of nutrient-limited environments. Our documentation includes three examples on how to modify the reward function, if needed, via the Gym reward wrappers. 
The agromanagement YAML file defines crop and soil dynamics and specifies the weather data which is provided by NASAPower. Gym environments, reward wrappers, and agromanagement files are \emph{configurable}, allowing customization to simulate agromanagement decision problems across various crops, farms, and tasks.

\paragraph{States and Observations}
The model state in WOFOSTGym is the concatenation of two feature vectors, $\mathbf{c}=(c_1\ldots c_{203})$ and $\mathbf{w}=(w_1,\ldots, w_7)$, where $\mathbf{c}$ contains the crop and soil state and $\mathbf{w}$ contains the weather state for a given day. However, most of these state features are not directly observable in the real-world. Thus, the state features available to the RL agent are a subset of the model state as observation $\mathbf{o}=(c_1,...,c_n)$, with $n\ll210$. An observation could be: $\mathbf{o}=$ (Weight of Storage Organs, Development Stage, Leaf Area Index, Soil Moisture Content, Rainfall, Solar Irradiation, Daily Temperature). WOFOSTGym supports any combination of state features as an observation. In the multi-farm environments, the agent receives an observation for each farm and the daily weather observation is shared across farms.

\paragraph{Action Space}
WOFOSTGym's action space consists of fertilization (F): nitrogen (n), phosphorus (p), and potassium (k), and irrigation (I) actions. At each time step, an action can be chosen from $A=\{F_n, F_p, F_k, I\}$ which corresponds to applying fertilizer ($F_i$) or water ($I$) in the following amounts, where $f$, $n$, $i$, and $m$ can all be modified:

\vspace{-5pt}
\begin{center}
\scalebox{1}{
       $ F_i=\left\{f\cdot k \frac{\text{kg}}{\text{ha}}\Big|k\in \{0\ldots n\}\right\}, I=\left\{i\cdot k \frac{\text{cm}}{\text{ha}}\Big|k\in \{0\ldots m\}\right\} $
}
\end{center}
\vspace{-5pt}

meaning that $|A|=3n+m$. By default, a time step represents a single day, but can be modified to denote multiple weeks to model the varying length between agromanagement decisions. 

\paragraph{Reward}
Real-world agriculture requires balancing yield with constraints such as fertilizer costs, water usage limits, and surface runoff restrictions. WOFOSTGym includes reward wrappers to penalize the violation of these constraints. By default, positive reward is a function of crop yield, as profitability is the primary driver of agromanagement policy adoption \citep{turchetta_learning_2022}. However, to enable wide configurability, WOFOSTGym's reward wrapper design enables the reward to be any function of the \emph{entire state space}. An example reward function in WOFOSTGym is: $ R_t = \text{Yield} -C\cdot(F_t+I_t)$, where $C$ is a constant that modifies the penalty for nutrient application. 

\paragraph{Domain Randomization} Domain randomization enables successful sim-to-real transfer~\citep{mehta_active_2020}, a key feature missing from existing crop simulators. WOFOSTGym supports three types of domain randomization, which can be used individually, in combination, or not at all. They are: 1) adding small amounts of random uniform noise to parameters in the WOFOST GGM, 2) allowing RL agents to train on different crops and soil types simultaneously, and 3) enabling RL agents to train on a wide breadth of historical weather data.

\paragraph{Available Crops and Modifications to WOFOST} WOFOSTGym includes parameters for 23 annual crops and 2 perennial crops which were all calibrated empirically from field data~\citep{de_wit_ajwdewitwofost_crop_parameters_2025,wang_pear_2022, bai_assessing_2019}. For perennial crops, it is insufficient to model individual seasons of crop growth, as important agromanagement decisions are made during the dormant season~\citep{forcella_real-time_1998}. It is more appropriate to model growth over multiple consecutive years, which requires modification to the phenology, crop organ, and nutrient balance modules WOFOST. We outline the modifications made to WOFOST and list all available crops in the Appendix. 

\subsection{Parameter Calibration for Crop Growth Models}
Before a CGM can be used for sim-to-real transfer with RL, it must be calibrated~\citep{bhatia_crop_2014}. CGM parameters are typically derived from field experiments and optimized using regression to find the best fit~\citep{berghuijs_expanding_2024,zapata_predicting_2017}. Parameter spaces for CGMs are high-dimensional and highly non-linear~\citep{sinclair_criteria_2000}, so brute force and regression techniques that are commonly used in agronomy research may be insufficient to find an optimal solution. To overcome the limitations in current CGM calibration methods, we propose a Bayesian optimization approach that requires minimal domain knowledge and  outperforms the regression-based methods. When historical crop data is available, Bayesian optimization is a more principled way of exploring the parameter space to increase the model fidelity of WOFOSTGym.

\paragraph{Example: Bayesian Optimization for Grape Phenology Calibration}
\label{sec:grape_bo}
Grape phenology is divided into three key phenological stages: Bud Break, Bloom, and Veraison~\citep{lorenz_growth_1995}. Accurately predicting the onset of a phenological stage allows growers to implement effective agromanagement strategies, and the Root Mean Squared Error (RMSE) is the widely accepted measure of performance in grape phenology modeling~\citep{parker_classification_2013}. We use an \emph{iterative} optimization process that uses Bayesian optimization in each iteration to refine parameters and minimize error across \emph{all} phenological stages. Phenology in WOFOSTGym is described by a set of seven parameters, $\theta$. Each iteration aligns with minimizing RMSE for a stage $k$, where $\theta_k$ is a subset of $\theta$.

Using a dataset of six to 15 years of historical weather and phenology observations per cultivar collected by~\citet{zapata_predicting_2017}, we define the following loss function for Bayesian Optimization: 

\vspace{-5pt}
\begin{center}
\scalebox{1}{
       $ L_{RMSE}(\theta_k) = \sqrt{\frac 1n\overset{n}{\underset{i=1}{\sum}}\left(P_i^k(\theta_k)-O_i^k\right)^2+\frac1n\overset{n}{\underset{i=1}{\sum}}\left(P_i^{k-1}(\theta_k)-O_i^{k-1}\right)^2}$
}
\end{center}
\vspace{-5pt}

where $P_i^k(\theta_k)$ and $O_i^k$ denote the predicted and observed onset day for phenological stage $k$ for year $i$ with parameter set $\theta_k$. We run three iterations of Bayesian optimization~\citep{noguiera_bayesian_2014} with a RBF kernel and the expected improvement acquisition function for 500 steps. By retaining the best-fit parameters found by each iteration, we obtain $\theta=\{\theta_\text{Bud Break},\theta_\text{Bloom},\theta_\text{Veraison}\}$, which minimizes the RMSE across all phenological stages. We compare our Bayesian Optimization results with~\citet{zapata_predicting_2017} who use linear regression. They find parameter sets for grape phenology, BB-$T_b$ and BL-$T_b$, that aim to minimize the error for Bud Break and Bloom, and report the RMSE for all stages. Our results in Table~\ref{tab:grape_bo_avg} show that our model outperforms others, providing a 10\% reduction in RMSE over the next best parameter set, BB-$T_b$.

\begin{table}[t]
\centering
\resizebox{\textwidth}{!}{%
\begin{tabular}{llll|lll|lll|lll}
\toprule
\textbf{Cultivar} & \multicolumn{3}{c}{\textbf{Bud Break}} & \multicolumn{3}{c}{\textbf{Bloom}} & \multicolumn{3}{c}{\textbf{Veraison}} & \multicolumn{3}{c}{\textbf{Cumulative Error}} \\
                     & Ours  & BB-$T_b$  & BL-$T_b$   & Ours & BB-$T_b$  & BL-$T_b$   & Ours & BB-$T_b$  & BL-$T_b$  & Ours  & BB-$T_b$  & BL-$T_b$ \\
\midrule
Cabernet Franc        & 4.0  & 6.1 & 6.2  & 3.5 & 3.1 & 2.9  & 7.7  & 6.7  & 7.1  & 15.2  & 15.9 & 16.2  \\
Cabernet Sauvignon    & 5.0  & 8.7 & 10.5 & 5.2 & 5.8 & 5.7  & 9.8  & 6.6  & 7.0  & 20.0  & 21.1 & 23.2  \\
Malbec                & 3.7  & 5.6 & 6.2  & 2.8 & 3.2 & 2.9  & 8.3  & 5.7  & 6.0  & 14.8  & 14.5 & 15.1  \\
Pinot Noir            & 3.6  & 4.2 & 3.9  & 2.4 & 2.6 & 2.3  & 8.6  & 6.6  & 7.7  & 14.6  & 13.4 & 13.9  \\
Zinfandel             & 3.7  & 6.8 & 9.0  & 3.8 & 4.3 & 4.0  & 6.0  & 4.1  & 3.8  & 13.5  & 15.2 & 16.8  \\
Chardonnay            & 7.2  & 6.3 & 5.9  & 4.1 & 3.7 & 3.2  & 7.8  & 5.6  & 5.9  & 19.1  & 15.6 & 15.0  \\
Chenin Blanc          & 5.0  & 6.1 & 6.2  & 3.8 & 4.8 & 4.6  & 8.5  & 9.2  & 9.4  & 17.3  & 20.1 & 20.2  \\
Sauvignon Blanc       & 3.4  & 6.4 & 5.7  & 5.9 & 3.7 & 3.5  & 1.6  & 7.7  & 8.5  & 10.9  & 17.8 & 17.7  \\
Semillon              & 4.7  & 6.0 & 7.0  & 2.7 & 6.0 & 5.8  & 8.8  & 11.2 & 11.6 & 16.2  & 23.2 & 24.4  \\
Riesling              & 3.7  & 4.2 & 5.7  & 3.8 & 4.1 & 3.7  & 8.5  & 8.5  & 9.0  & 16.0  & 16.8 & 18.4  \\
\midrule
\textbf{Average}      & 4.4  & 6.0 & 6.6  & 3.8 & 4.1 & 3.9  & 7.4  & 7.2  & 7.6  & 15.6  & 17.3 & 18.1  \\
\bottomrule
\end{tabular}%
}
\caption{RMSE in days when predicting the key phenological stages (Bud Break, Bloom, and Veraison) in ten grape cultivars. The columns represent the RMSE between the model's predicted phenology for a given parameterization and the observed phenology. Ours: Using parameter set tuned with Bayesian Optimization. BB-$T_b$: Parameter set tuned for Bud Break. BL-$T_b$: Parameter set for Bloom. Values for BB-$T_b$ and BL-$T_b$ are columns 2 and 3 in Table 6 in~\citet{zapata_predicting_2017}.}
\label{tab:grape_bo_avg}
\end{table}

The 32 calibrated grape phenology parameterizations included in WOFOSTGym increase model fidelity and represent a step towards sim-to-real transfer for crop management policies in open-field agriculture. Grape growers can use the high-fidelity phenology models in WOFOSTGym as a \emph{digital twin} to examine the effects of different agromanagement policies on their grape vines without the risk of crop loss. As more crop data becomes widely available, our Bayesian optimization method can be used to calibrate CGM parameters to more accurately model a variety of crop processes.

\section{Experiments and Results}
\label{sec:experiments}

To illustrate the use of WOFOSTGym, we run RL and IL experiments on diverse tasks to learn crop management policies for annual and perennial crops under realistic constraints. 
We present results using varied crops and soil types, demonstrating  WOFOSTGym's customizability. Overall, our results show that off-the-shelf RL algorithms struggle with hard constraints, long horizons, and delayed feedback---challenges inherent to agriculture and captured by WOFOSTGym, making it a valuable platform for both core RL research and agromanagement decision support.

Our crop selection was guided by common agronomic challenges: wheat and barley for nutrient-limited growth due to their high nitrogen and water demands, and potatoes for soil nutrient runoff risks. We also test with pears and jujubes for the long-horizon decision-making challenges in perennial crop management, and maize as it is the only crop supported by other simulators. 

Agents in our experiments can choose from 16 actions, each corresponding to one of four discrete amounts of nitrogen, phosphorus, potassium, and water. Unless otherwise noted, the agent observes the state features: development stage; weight of storage organs; total nitrogen, phosphorus, potassium, and water applied; soil moisture content; nitrogen, phosphorus, and potassium in subsoil; solar irradiation; average temperature; and rainfall. For our RL experiments, we use PPO, SAC, and DQN~\citep{schulman_proximal_2017, haarnoja_soft_2018, mnih_playing_2013}, using implementations from~\citet{huang_cleanrl_2021} and  
hyperparameters tuned experimentally to yield best performance in the WOFOSTGym domain. For our IL experiments, we use implementations from~\citet{gleave_imitation_2022} of BC, GAIL, and AIRL~\citep{furukawa_framework_2000,ho_generative_2016,fu_learning_2018}. GAIL and AIRL use a PPO policy, and BC uses an Actor Critic Policy, all written by~\citet{raffin_stable-baselines3_2021}. All experiments and code can be found at: \url{https://github.com/Intelligent-Reliable-Autonomous-Systems/WOFOSTGym}.

\begin{figure}[t]
    \centering
    \includegraphics[width=\textwidth]{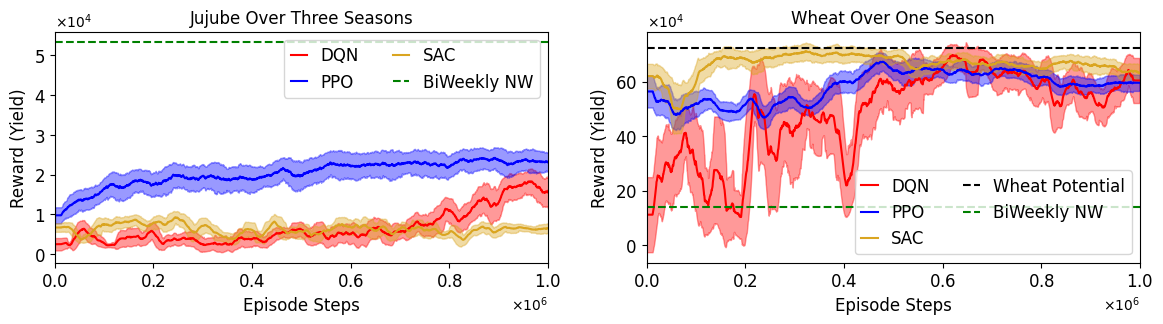}
    \caption{Unconstrained Control. The average reward, as seasonal yield, of different policies. The BiWeekly NW policy alternates applying nitrogen and water biweekly while the Wheat Potential is the maximum growth obtainable. We omit the Jujube Potential because it assumes daily intervention, while we only allow biweekly intervention.}
    \label{fig:baseline_agents}
    
\end{figure}

\paragraph{Learning Efficiency}

Figure \ref{fig:baseline_agents} presents learning curves for maximizing jujube growth over three seasons and wheat over one season in WOFOSTGym. We compare RL performance against the maximum potential yield and an agromanagement policy that alternates nitrogen fertilization and irrigation biweekly. In the wheat experiment, the RL algorithms significantly outperform the baseline of a bi-weekly nitrogen and water application policy but fall short of reaching the potential production of an unlimited nutrient setting. For jujube, we see that RL agents are unable to match the performance of a monthly nitrogen and water application policy. These examples show the potential for off-the-shelf RL algorithms to achieve non-trivial performance for some crop scenarios, but also indicate that there is significant room for improvement in others.

\begin{figure}[h]
    \centering
    \includegraphics[width=\textwidth]{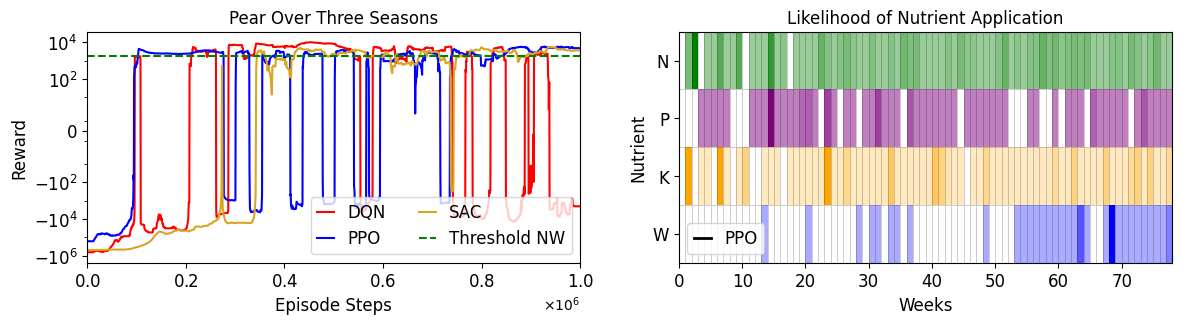}
    \caption{Constrained Control. (Left) The running average of the reward during training. (Right) The likelihood of fertilization or irrigation action each week. Likelihoods were computed over 30 episodes with darker colors signifying more likely nutrient application.}
    \label{fig:threshold_agents}  
\end{figure}

\paragraph{Learning Under Constraints}
In the real world, yield maximization is always subject to multiple constraints, such as a limit on the amount of fertilizer and water that can be applied per season. To model this, we reward total yield and apply a large negative penalty if the fertilization and irrigation thresholds (in kg/ha and cm/ha) are exceeded. Figure \ref{fig:threshold_agents} shows the results of RL algorithms with this reward function; positive reward indicates no constraint violation, rewards less than zero indicate a constraint violation, and rewards less than $-10^5$ indicate more than 5 constraint violations. Note that unlike the previous experiment, there is no principled way to find the maximum reward obtainable in this setting. We compare the RL agents to a baseline that applies nitrogen fertilizer and water until it meets the same thresholds of 80 kg/ha of fertilizer or 40 cm/ha of water~\citep{bushong_evaluation_2016}. While this baseline satisfies constraints, it achieves a lower average reward than the trained RL agents. It is clear that this simple approach to handling constraints is insufficient for such real-world crop scenarios, which shows WOFOSTGym's potential as a testbed for constrained RL research. 

\begin{figure}[t]
    \centering
    \includegraphics[width=\textwidth]{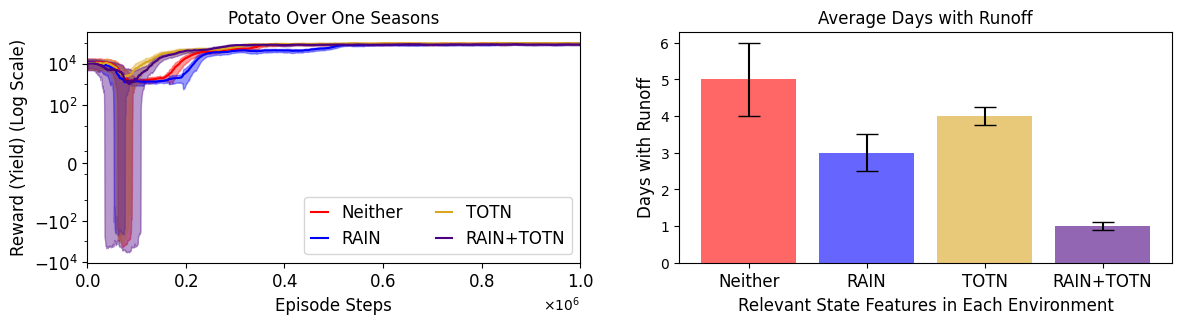}
    \caption{Constrained Control Under Partial Observability. The average reward of PPO agents during training and the average days of runoff after completing training over 15 seasons. }
    \label{fig:fert_limited_agents}
\end{figure}

\paragraph{Effect of Partial Observability on Constraint Adherence}
\label{sec:rl_runoff}
Limitations on sensing capabilities are a constant source of uncertainty in agromanagement decisions. To illustrate the effects of partial observability in WOFOSTGym, we consider two relevant state features, RAIN and TOTN, the daily rainfall and fertilizer on the soil surface, respectively, and create four partially observable environments based on the omission of the two variables from the observation space. We design a reward function that rewards crop yield subject to a $-10^4$ penalty if nutrient runoff occurs. Nutrient runoff happens when fertilizer amasses on the soil surface and irrigation or rainfall occur.  We train a PPO agent to grow the potato crop in each environment and show the results in Figure \ref{fig:fert_limited_agents}. Access to all relevant features improves constraint adherence of an agent policy. However, even in the fully observable case, constraint satisfaction is not guaranteed, exhibited by the non-zero days of runoff on average. Future research could use WOFOSTGym to study constraint adherence in partially observable environments, and also inform the importance of obtaining costly field measurements.

\paragraph{Imitation Learning for Agromanagement Decisions} 
IL has been proposed as a method for sim-to-real transfer in agriculture, by learning from  demonstrations of a trained RL agent~\citep{tao_optimizing_2023}. We investigate the ability of BC, GAIL, and AIRL to learn from demonstrations in WOFOSTGym. We provide each IL agent with 100 seasons of data generated from an expert PPO agent trained to maximize barley yield subject to strict limits of 20 kg/ha of fertilizer and 20 cm/ha of water per season. Our results in Table \ref{tab:il_agents} show that GAIL and AIRL failed to obtain similar yield to the expert and also exhibited very different behavior as shown by the differences in nitrogen and water application. Surprisingly, BC demonstrated the closest matching behavior, although it was unable to fully avoid constraint violations. This shows that WOFOSTGym can serve as a non-trivial benchmark for IL and in particular, research on implementing constraints into IL.  

\begin{table}[h]
    \centering
\resizebox{\textwidth}{!}{%
    \begin{tabular}{lllllll}
        \toprule
        Agent & \makecell{Max Yield\\(kg/ha)} & \makecell{Constraints\\Violated} & \makecell{Nitrogen\\(kg/ha)} & \makecell{Phosphorus\\(kg/ha)} & \makecell{Potassium\\(kg/ha)} & \makecell{Water\\(cm/ha)}\\
        \midrule
        Expert (PPO) & $4376 \pm 805$ & $0.00 \pm 0.00$ & $16.53 \pm 3.3$ & $6.13 \pm 2.47$ & $14.53 \pm 3.46$ & $3.33 \pm 0.83$ \\
        AIRL & $2975 \pm 335$ & $4.53 \pm 2.33$ & $28.93 \pm 5.56$ & $2.67 \pm 2.02$ & $4.40 \pm 2.65$ & $1.80 \pm 0.93$ \\
        GAIL & $2647 \pm 562$ & $0.67 \pm 0.94$ & $8.93 \pm 2.72$ & $9.47 \pm 4.10$ & $19.87 \pm 4.29$ & $4.00 \pm 1.13$ \\
        BC & $4598 \pm 790$ & $0.33 \pm 0.6$ & $16.8 \pm 3.56$ & $6.53 \pm 3.38$ & $14.67 \pm 2.98$ & $4.3 \pm 1.25$ \\

        \bottomrule
    \end{tabular}%
    }
    \caption{Results of three IL agents trained to maximize barley yield subject to constraints on nutrient application. The Constraints Violated column shows the number of days where excess nutrients were applied after the threshold was reached. Results are averaged over 15 seasons.}
    \label{tab:il_agents}
    
\end{table}

\paragraph{Comparison of Agromanagement Decisions on Multiple Farms}
Comparing yield and nutrient levels under different agromanagement policies is desirable for farmers, but unrealistic to perform in the field due to the risk of exploratory actions decreasing crop yield. WOFOSTGym enables agromanagement policies to be evaluated in simulation which could be a useful tool for farmers to understand the impacts of agromanagement decisions on crop and soil health. WOFOSTGym instances describe the dynamics of a field growing a specific crop. Each field can represent a farm. Using WOFOSTGym, we compare joint multi-field policies with field-specific policies to analyze their trade-offs in accumulated yield.

\begin{figure}[t]
    \centering
    \includegraphics[width=\linewidth]{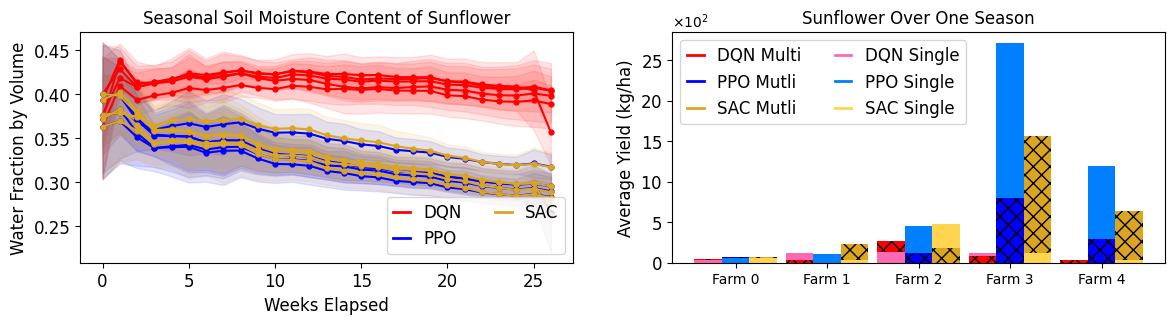}
    \caption{(Left) The soil moisture content of each field under three joint RL agromanagement policies. (Right) The average yield obtained by trained multi-field agents. Lighter colors indicate the yield obtained by an agent trained on that specific field as a baseline for obtainable crop yield.}
    \label{fig:multi_farm}
\end{figure}

As farmers often apply the same policy to multiple fields, we create a WOFOSTGym environment that simulates the growth of five sunflower fields experiencing the same weather. The observation space is the growth and soil variables for each field. The weather is shared between fields and the action selected is uniformly applied to each field. We train a PPO, DQN, and SAC agent in this multi-field scenario and report the soil moisture content and the average yield with each agent policy on each field in Figure \ref{fig:multi_farm}. We then train the respective agents on each individual field to understand the value of using a specialized policy compared to a joint policy. The increased soil moisture content achieved under the DQN policy leads to the lowest yield across all fields, providing valuable insight into soil dynamics.

\paragraph{Simulator Run Times}

Fast simulators are central to the successful application of RL given the high sampling complexity of RL algorithms~\citep{lechner_gigastep_2023}. We benchmark the run times of three crop simulators: WOFOSTGym, CyclesGym, and gym-DSSAT~\citep{gautron_gym-dssat_2022, turchetta_learning_2022}. We compare run times for a single episode of growing the maize crop (155 episode steps). Given the large potential overhead when resetting the underlying CGM, we also measure the run times of the step and reset functions.
Our results in Table \ref{tab:times} show that WOFOSTGym outperforms CyclesGym, the only crop simulator that supports multi-year simulations, by an order of magnitude. WOFOSTGym is also faster than gym-DSSAT, due to its significantly faster reset function. 
Although gym-DSSAT has a faster step function, WOFOSTGym performs more computations per step by maintaining nitrogen, phosphorus, and potassium balances, whereas gym-DSSAT maintains only a nitrogen balance. 

\begin{table}[h]
    \centering
    \begin{tabular}{llll}
    \toprule
     Run Time (s)   & WOFOSTGym & CyclesGym & gym-DSSAT \\
    \midrule
     1 Episode & 0.34$\pm0.012$ & 2.08$\pm$.221 & 0.38$\pm$0.018\\
     Step Function & 0.003$\pm$0.0005 & 0.04$\pm$.0020& 0.001$\pm$0.0001 \\
     Reset Function &  0.012$\pm$0.002 & 0.055$\pm$.002 & 0.191$\pm$0.012 \\
    \bottomrule
    \end{tabular}
    \caption{The average runtime and standard deviation, computed over 100 trials, of three different crop simulators on an Nvidia 3080Ti.}

    \label{tab:times}
    
\end{table}
\section{Limitations and Future Work}
WOFOSTGym takes around two seconds to run a three-year simulation of a perennial jujube crop. Although WOFOSTGym offers an improved run time compared to other crop simulators, the run time quickly adds up when RL algorithms require millions of episodes to learn a good policy. As episode horizon increases for modeling perennial crop management decisions, accelerating the modeling of crop dynamics will become crucial.

WOFOSTGym is designed for modeling crop growth of a specific crop and currently does not support optimizing long-term crop rotation strategies. As the WOFOST CGM supports crop rotations, extending WOFOSTGym to support such problems is a promising extension. 

Although the parameter sets in WOFOSTGym can be considered high-fidelity models as they were tuned against field data, sim-to-real transfer using WOFOSTGym should be attempted with caution. As research bridging RL to open-field agriculture advances and CGM fidelity improves through approaches like those in Section \ref{sec:grape_bo}, direct sim-to-real transfer may become increasingly feasible.  

\section{Summary}
\label{sec:conclusion}
We present WOFOSTGym, the first RL simulator for annual and perennial crop management decision support. The WOFOSTGym repository includes high-fidelity parameters for two perennial crops and 23 annual crops, along with diverse pre-specified agromanagement policies for benchmarking RL agents. Its customizable design enables researchers to conduct experiments without requiring agricultural domain expertise. To improve CGM fidelity and facilitate sim-to-real transfer in open-field agriculture, we propose a Bayesian optimization-based calibration method. Our results reveal the limitations of current RL and IL algorithms in this domain, emphasizing the need for further research to address the specific challenges presented in the agriculture domain. We outline realistic benchmarks to assess RL algorithms before deployment for agricultural decision support.

\subsubsection*{Broader Impact Statement}
\label{sec:broaderImpact}
Reinforcement learning for crop modeling has the potential to help growers optimize yield while reducing costs and environmental impact. WOFOSTGym provides a high-fidelity platform for researchers to develop and evaluate agromanagement policies. However, due to the gap between simulation and real-world environments, RL policy performance in simulation may not translate directly to field trials.

\subsubsection*{Acknowledgments}
 \label{sec:ack}
 This research was supported by USDA NIFA award No.
2021-67021-35344 (AgAID AI Institute).

\newpage 
\bibliography{references}
\bibliographystyle{rlj}

\newpage
\appendix

\section{WOFOSTGym Modified Perennial Crop Growth Model}
\label{supp:wofost_perennial}

Prior works have shown that WOFOST CGM could be modified to model the growth of pear and jujube crops across multiple growing seasons~\citet{de_wit_pcse_2024,bai_assessing_2019, wang_pear_2022}, establishing it as a solid foundation for developing a perennial crop simulator. 
Below, we outline the key modifications we made to WOFOST to support perennial crop modeling.

\paragraph{Perennial Phenology} We primarily focused on modifying the phenology submodule within the WOFOST CGM to account for the differences between annual and perennial crop phenology. Unlike annual crops, the phenology of perennial crops is characterized by a dormancy stage induced by day length in autumn and released by temperature in spring~\citep{rohde_plant_2007}. To capture this behavior, we introduced parameters for dormancy induction based on day length, release temperature threshold, and minimum dormancy duration. In our modified WOFOST CGM, dormancy can also be triggered by prolonged growth stagnation, indicating insufficient ambient temperature for crop growth~\citep{jones_pattern_1978}.

\begin{figure}[h]
    \centering
    \includegraphics[width=0.98\linewidth]{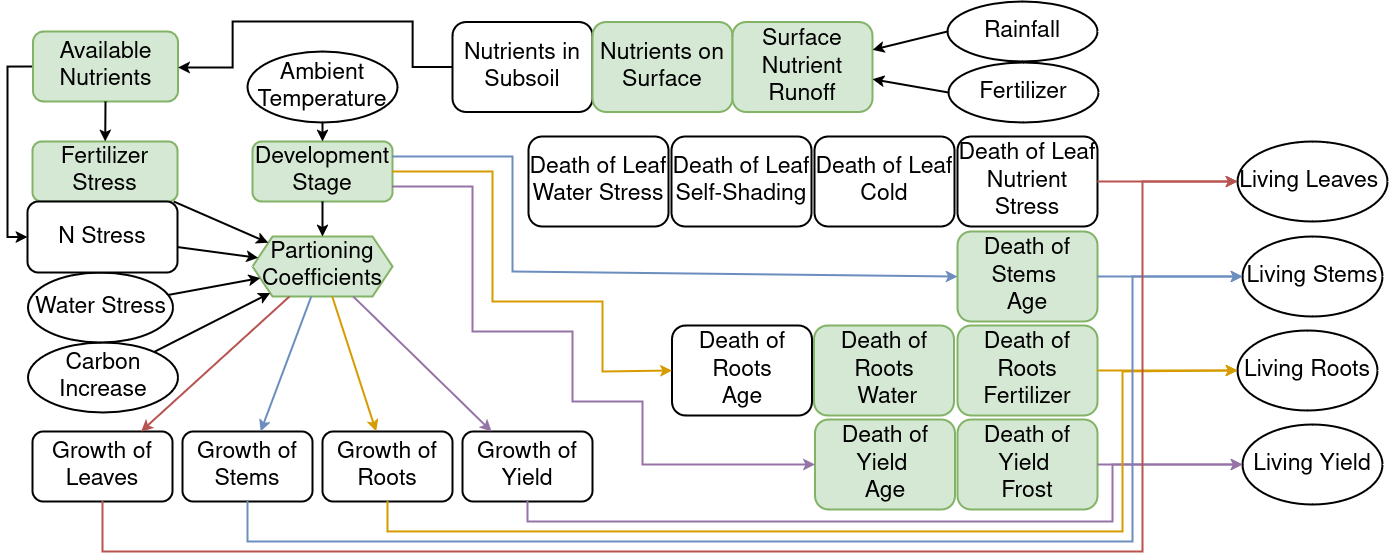}
    \caption{A simplified flowchart of the perennial crop growth in WOFOSTGym. Boxes highlighted in green denote additions or areas of substantial change to the underlying WOFOST CGM to support perennial crop growth. The development stage of the crop is driven by the daily ambient temperature. The development stage determines how accumulated dry matter is partitioned to crop organs subject to available nutrients. The weight of the living organs (yield) is calculated as the accumulated difference between the growth and death rates.}
    \label{fig:dvs_growth}
\end{figure}

\paragraph{Perennial Organ Growth}
In addition to differences in phenology, perennial crops exhibit differences in their visible growth organs~\citep{thomas_annuality_2000}. The roots and stems of perennial crops survive year-round, while the leaves and storage organs regrow each season subject to intercepted light and nutrient uptake. Crop organ death rates are modeled as a function of the development stage of the crop~\citep{linden_logit_1996}. Notably, perennials exhibit a reduced seasonal growth as they age~\citep{munne-bosch_aging_2007}. 
While the underlying mechanisms for reduced growth remain difficult to quantify, we model this decline empirically through increased maintenance respiration and decreased carbon conversion efficiency as a function of age~\citep{zhu_developing_2021}. See Figure \ref{fig:dvs_growth} for a visual overview of how key features evolve throughout the course of a perennial crop simulation.

\paragraph{Modified Nutrient Module}
A multi-layer nutrient balance is important for modeling the effects of fertilization stressors on the roots, stems, and nutrient partitioning~\citep{albornoz_crop_2016}. We extend WOFOST's single-layer nutrient balance to a multi-layered nutrient balance within the soil module~\citep{he_modeling_2013}. When nutrients are applied via fertilization, they reside on the soil surface. As the simulation evolves, nutrients are absorbed into the subsoil and then the roots of the plant. When surface nutrient levels are too high, the partitioning of dry matter is changed to limit allocation to the storage organs in favor of stems and leaves~\citep{he_modeling_2013}.

\section{Available Crops} 
\label{supp:available_crops}
WOFOSTGym includes parameters for two perennial crops: pear and jujube,
and 23 annual crops: barley, cassava, chickpea, cotton, cowpea, faba bean, groundnut, maize, millet, mung bean, pigeon pea, potato, rapeseed, rice, onion, sorghum, soybean, sugar beet, sugarcane, sunflower, sweet potato, tobacco, and wheat. Each crop contains between \emph{one and ten varieties}. WOFOST CGM parameters for each variety were calibrated empirically from field data~\citep{de_wit_ajwdewitwofost_crop_parameters_2025}. By modeling each crop variety as a task, agromanagement decisions for multiple crop varieties can be optimized with multi-task RL~\citep{hessel_multi-task_2019}.

In addition to the high-fidelity models for 25 crops, WOFOSTGym also includes parameters for modeling the phenology of 32 grape cultivars. These cultivars are: Aligote, Alvarinho, Auxerrois, Barbera, Cabernet Franc, Cabernet Sauvignon, Chardonnay, Chenin Blanc, Concord, Dolcetto, Durif, Gewurztraminer, Green Veltliner, Grenache, Lemberger, Malbec, Melon, Merlot, Muscat Blanc, Nebbiolo, Petit Verdot, Pinot Blanc, Pinot Gris, Pinot Noir, Riesling, Sangiovese, Sauvignon Blanc, Semillon, Syrah, Tempranillo, Viognier, and Zinfandel.

\section{Crop Growth Models}
\label{supp:cgm}
CGMs are typically one of three types: mechanistic, empirical, or hybrid. Mechanistic models simulate canopy or nutrient level crop processes to validate scientific understanding of crop growth~\citep{estes_comparing_2013}. Empirical models rely on observed field data, offering greater scalability with lower computational overhead~\citep{di_paola_overview_2016}. Hybrid crop models simulate crop growth using both mechanistic and empirical modeling decisions~\citep{yang_hybrid-maizemaize_2004}. 

WOFOST is a single-year and multi-crop agroecosystem model~\citep{jones_brief_2017}. It relies both on mechanistic and empirical processes to simulate crop growth~\citep{di_paola_overview_2016}. Crop growth in WOFOST is determined by the atmospheric CO2 concentration, irradiation, daily temperature, subject to limited water, nitrogen, phosphorus, and potassium. While WOFOST was designed for simulating the yield of annual crops~\citep{van_diepen_wofost_1989}, field studies have shown that it can be used to accurately predict yield in perennial fruit trees with some small modifications to the base model~\citet{wang_pear_2022, bai_assessing_2019}

Given WOFOST's ability to simulate a wide variety of crop and soil dynamics, and its modular implementation in Python, WOFOST is an ideal CGM candidate to be used to simulate perennial crop growth to address that lack of perennial CGMs available, and the lack of perennial crop benchmarks available for RL research~\citep{gautron_reinforcement_2022}. For an introduction to the WOFOST, we refer readers to the works of ~\citet{de_wit_pcse_2024} and~\citet{de_wit_wofost_2019}. There are a wide variety of CGMs available for use. Table \ref{tab:cgm} outlines the desiderata used to select WOFOST as the CGM for  WOFOSTGym.

\begin{table}[t]
    \centering
    \begin{tabular}{c c c c c c c}
    \hline
    Crop Model & Model Type  & Nutrient Balance & Water Balance & Crop Type &  Language \\ 
    \hline
    WOFOST    & Hybrid &  \makecell{nitrogen,\\phosphorus,\\potassium} & \makecell{Single Layer,\\Multi Layer} & Annual & \makecell{Python,\\FORTRAN} \\ 
    \hline
    APSIM     & Mechanistic   & \makecell{nitrogen,\\phosphorus,\\potassium} & Multi Layer & Annual & \makecell{FORTRAN,\\C++} \\
    \hline
    DSSAT     & Hybrid     & \makecell{nitrogen,\\phosphorus,\\potassium} & Multi Layer & Annual & \makecell{FORTRAN}\\
    \hline
    CropSyst & Mechanistic   & nitrogen & Single Layer & Annual & C++\\
    \hline
    EPIC      & Hybrid   & \makecell{nitrogen,\\phosphorus} & Multi Layer & \makecell{Annual,\\Rotations}&  FORTRAN\\
    \hline
    STICS     & Empirical  & nitrogen & Multi Layer & \makecell{Annual} & \makecell{Executable}\\ 
    \hline
    Cycles    & Mechanistic & nitrogen & Multi Layer & \makecell{Annual,\\Rotations} & Executable \\
    \hline
    AquaCrop  & Empirical & abundant & Multi Layer & Annual & \makecell{Python,\\Executable}\\
    \hline
    LINTUL3    & Empirical  & nitrogen & Abundant & Annual & \makecell{Python}\\
    \hline
    \end{tabular}
    \caption{Different CGMs and their strengths and weaknesses for modeling high fidelity crop growth, interfacing with RL algorithms, and supporting perennial crop decision evaluation.}
    \label{tab:cgm}
\end{table}

\section{Configurability of WOFOSTGym}
\label{append:usability}

Other crop simulators are difficult for RL researchers to use because of their unfamiliarity with CGMs. WOFOSTGym aims to relieve the burden of domain knowledge required to use other crop simulators by streamlining configuration into readable YAML files. In this section, we highlight the features that make WOFOSTGym easy to use for RL researchers interested in agriculture. 

\paragraph{Simulation Configuration} 
WOFOST CGM configuration is divided into three configuration files: the crop YAML file, site YAML file, and agromanagement YAML file. WOFOSTGym provides 25 crop YAML files and three site YAML files. In the agromanagement YAML file, the specific crop and site configuration is specified, along with the length of the simulation, year, and geographic location. The agromanagement YAML file contains 14 entries, while still enabling a user to simulate 25 different crops with one to ten varieties per crop. This feature is an improvement over other crop simulators which support only 1 crop. 

With domain knowledge, a crop or site can be added by calibrating the parameters in the crop and site YAML files against real-world data. Finally, every parameter can be modified from the command line and each simulation saves a configuration file, aiding reproducibility (see Figure \ref{fig:wofost_demo}).

\begin{figure}[h]
\begin{lstlisting}[language=bash]
# Test Simulation of the Jujube crop
python3 test_wofost.py --save-folder test/ --data-file test --npk.ag.crop-name jujube --npk.ag.crop-variety jujube_1 --env-id perennial-lnpkw-v0

# Generate data of the default crop (wheat) and modify a few crop parameters
python3 gen_data.py --save-folder data/ --data-file wheat_data --file-type npz --npk.wf.TBASEM 2.0 --npk.wf.SMFCF 0.51

# Train a SAC agent to irrigate the wheat crop in an environment where nitrogen, phosphorus and potassium nutrients are abundant and modify the SAC algorithm parameters
python3 train_agent.py --save-folder RL/ --agent-type SAC --env-id lnw-v0 --SAC.gamma 0.95

# Train a PPO agent given a WOFOSTGym Configuration file
python3 train_agent.py --save-folder RL --agent-type PPO --config-fpath RL/ppo_test.yaml
\end{lstlisting}
\caption{Example for how to configure agent training and data generation in WOFOSTGym. Specific parameters can be modified by the command line. In addition, configuration YAML files can also be loaded for reproducibility (and are automatically saved each time a simulation is run).}
\label{fig:wofost_demo}
\end{figure}

\paragraph{Pre-Specified Agricultural Policies}
The goal of using RL for agriculture is to improve upon crop management strategies. However, to measure this improvement, common crop management policies must be available in a crop simulator to compare against. While other crop simulators do not include baselines, WOFOSTGym comes with 10 pre-specified agromanagement policies that are commonly used in agriculture. These policies include: "fertilize $X$ amount every week," or "irrigate $X$ amount if the soil moisture content is below $Y$." In WOFOSTGym, $X$ and $Y$ are easily modifiable via command line or YAML file.  

\paragraph{Data Generation} Offline data is used to for problems in Offline RL~\citep{levine_offline_2020}, Off Policy Evaluation~\citep{thomas_data-efficient_2016}, and Transfer Learning~\citep{zhuang_comprehensive_2021}. To facilitate research on these topics in the agriculture domain, data must be widely available. However, available agricultural data is inaccessible or requires substantial preprocessing to account for missing data. To address this problem, crop simulators are a promising direction, yet no other available crop simulator provides an efficient pipeline for data generation. 

WOFOSTGym addresses this shortcoming by providing a \emph{pipeline} that can generate data from a variety of different crops and sites and supports generating data from both RL agent policies and the pre-specified policies described above. By including this data generation in WOFOSTGym, we hope to both facilitate research into these interesting RL related problems in agriculture, and set a standard of usability for crop simulators that follow WOFOSTGym.

\end{document}